\documentclass[runningheads]{llncs}
\usepackage{graphicx}
\usepackage{wrapfig}
\usepackage{caption}
\usepackage[caption=false]{subfig}
\usepackage{xcolor}
\usepackage{amssymb}
\usepackage{amsmath}
\usepackage[section]{placeins}
\usepackage{algpseudocode}
\usepackage{algorithm}
\usepackage[hidelinks]{hyperref}

\DeclareMathOperator*{\argmax}{arg\,max}

\newif\ifdraft
\drafttrue

\definecolor{orange}{rgb}{1,0.5,0}
\definecolor{pink}{rgb}{0.98, 0.38, 0.5}

\ifdraft
 \newcommand{\RS}[1]{{\color{red}{\bf RS: #1}}}

 \newcommand{\PMN}[1]{{\color{orange}{\bf PMN: #1}}}

 \newcommand{\MH}[1]{{\color{pink}{\bf MH: #1}}}

\else
 \newcommand{\RS}[1]{{\color{red}{}}}
 
 \newcommand{\PMN}[1]{{\color{red}{}}}
 
  \newcommand{\MH}[1]{{\color{red}{}}}
 
\fi

\newcommand{\comment}[1]{}

\newcommand{\eg}{e.\,g.,\ }
\newcommand{\ie}{i.\,e.,\ }

\begin{document}
\title{Concept-Centric Visual Turing Tests \\for Method Validation}
\titlerunning{Concept-Centric VTT}

\author{Tatiana Fountoukidou*\orcidID{0000-0001-9771-9609} \and
	Raphael Sznitman}
% index{Fountoukidou, Tatiana}
% index{Sznitman, Raphael}

%
\authorrunning{T. Fountoukidou et al.}

%
\institute{ARTORG Center, University of Bern, Bern, Switzerland\\
\email{\{tatiana.fountoukidou, raphael.sznitman\}@artorg.unibe.ch}}

\maketitle              % typeset the header of the contribution
\begin{abstract}
Recent advances in machine learning for medical imaging have led to impressive increases in model complexity and overall capabilities. However, the ability to discern the precise information a machine learning method is using to make decisions has lagged behind and it is often unclear how these performances are in fact achieved. Conventional evaluation metrics that reduce method performance to a single number or a curve only provide limited insights. Yet, systems used in clinical practice demand thorough validation that such crude characterizations miss. To this end, we present a framework to evaluate classification methods based on a number of interpretable concepts that are crucial for a clinical task. Our approach is inspired by the \emph{Turing Test} concept and how to devise a test that adaptively questions a method for its ability to interpret medical images. To do this, we make use of a Twenty Questions paradigm whereby we use a probabilistic model to characterize the method's capacity to grasp task-specific concepts, and we introduce a strategy to sequentially query the method according to its previous answers. The results show that the probabilistic model is able to expose both the dataset's and the method's biases, and can be used to reduce the number of queries needed for confident performance evaluation.
\end{abstract}

\section{Introduction}

The field of medical image computing (MIC) has radically changed with the emergence of large neural networks, or Deep Learning (DL). For MIC tasks that were long considered extremely challenging, such as image-based pathology classification and segmentation, DL methods have now reached human-level performances on a variety of benchmarks. 

Yet, as these methods have become increasingly powerful, the overall methodology to validate them has largely remained intact. For instance, challenge competitions compare different methods on a common dataset by using metrics most often borrowed from the computer vision literature. As recently noted in~\cite{MeierHein2019}, challenge competition rankings and outcomes are very often highly skewed to the dataset or metrics used, and rarely relate to the clinical task. To tackle this, recent developments in visual question answering (VQA)~\cite{Antol_2015_ICCV,ImageCLEFVQA-Med2018,lau2018dataset,wu2017visual} methods, which answer questions related to image content, show the ability to infer concepts beyond traditional classification. Here again, however, the metrics used to evaluate VQA's remain inadequate.

Instead, we consider an alternative approach to validating MIC methods, one inspired by Alan Turing's \emph{Turing Test} concept~\cite{machinery1950computing}, where a human unknowingly communicates either with another human or an Artificial Intelligence system that produces answers. The aim of the test is to distinguish between the two based on a set of asked questions. Turing tests have been used in medical imaging to evaluate the quality of adversarial attacks, by seeing if an expert can distinguish between a real and an adversarial example~\cite{chuquicusma2018fool,schlegl2019f}. Another approach, focusing on the interpretability of methods that infer semantic information from images (e.g. classification, segmentation etc.) is seen in the work of Geman et al.~\cite{geman2015visual} on automated Visual Turing Tests (VTT). In their work, an algorithm adaptively selects images and questions to pose to a method under evaluation (MuE) such that the answers can not be predicted from the history of answers. While this approach has increased explanatory power, it is limited to manually fabricated story lines to guide questioning. This makes it ill suited for medical applications where such story lines are hard to formalize.
\begin{figure}[!t]
	\centering
	\includegraphics[width=0.95\textwidth]{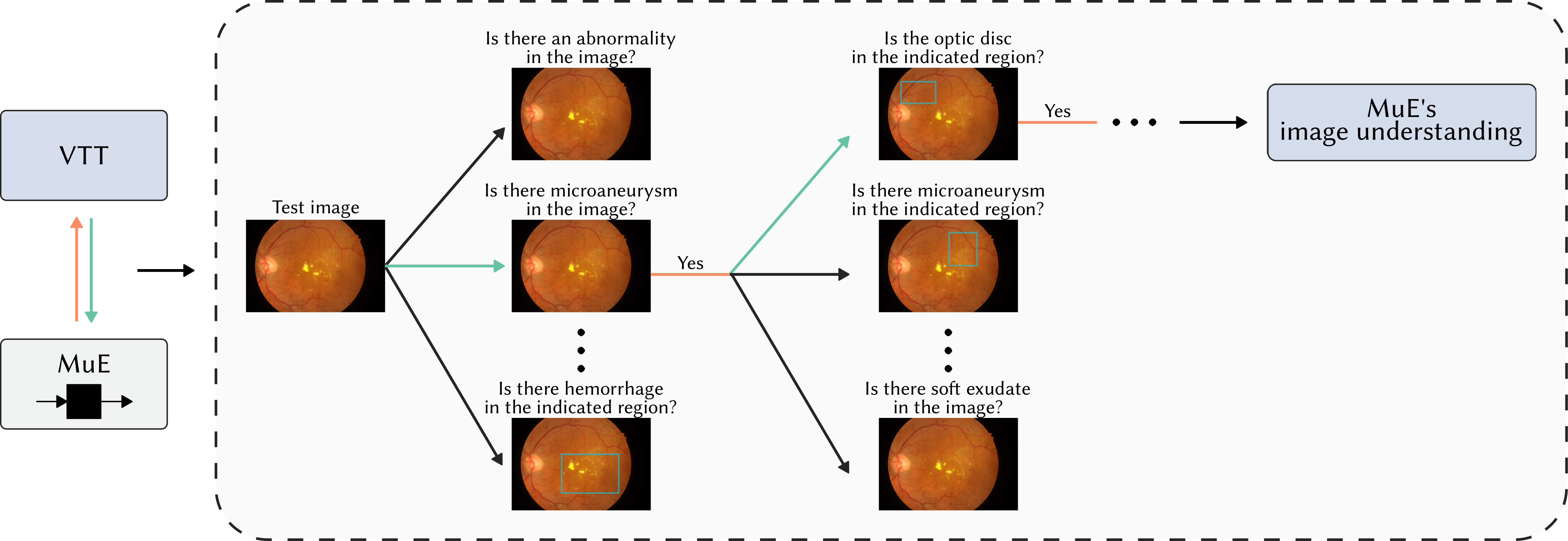}
	\caption{{Visual Turing Test (VTT) fundus image screening. Green arrows correspond to selected questions, and orange lines to the answers given by the MuE.}}
	\label{fig:VTT}
\end{figure}

For this reason, we propose a novel VTT framework to evaluate MIC methods (see Fig.~\ref{fig:VTT}). In particular, our approach focuses on evaluating MIC classification methods and we present a framework that tests if the method has correctly \emph{understood} the relevant medical concepts when inferring test data. We do this by formulating our problem as a Twenty Questions game~\cite{Bendig1953,Jedynak2012} in which we model the likelihood of a given MuE to provide correct answers for different concepts present in test images. Our framework then sequentially picks test images and concepts such that the uncertainty of this model is reduced as quickly as possible. We demonstrate our framework in the context of three different multi-label classification problems where each concept is encoded by a given binary label.

\section{Method}
Our proposed VTT framework evaluates how a MuE, a MIC classification method in this case, performs with respect to core concepts relevant to the task for which it was trained. To do this, we make use of a validation image dataset that the MuE has never had access to, $\mathcal{D} = \{\mathbf{s}_i\}_{i=1}^{N_S}$, where $\mathbf{s}_i$ is an available test sample, such as an image or an arbitrary region within an image. As such, $N_S$ can be excessively large, as potentially millions of regions can be extracted from a single test image. For each $\mathbf{s}_i$, we denote the potential concepts that could be present in the sample as, $\mathcal{C} = \{c_j\}_{j=1}^{N_C}$. From this, we define a ``question", $q = (\mathbf{s}_q, c_q) \in  \mathcal{D} \times \mathcal{C} = \mathcal{Q}$, of the form \textquotedblleft Is concept $c_q$ present in sample $\mathbf{s}_q$?\textquotedblright. We let $q_{gt} \in \{0,1\}$ be the true answer to question $q$. In this work, we consider MuEs that perform multi-label classification tasks, $f: \mathcal{Q} \rightarrow [0,1]$, taking as input the question $q$ and producing the probability of the answer being~\textquotedblleft Yes\textquotedblright (see Fig.~\ref{fig:VTT}). 

Given that evaluating all elements of $\mathcal{Q}$ may be computationally intractable, our VTT framework instead only evaluates a subset of $\mathcal{Q}$. We do this iteratively and adaptively, where we use the history of previously asked questions and their answers to build a {\it performance model} (Sec.~\ref{sec:perfModel}). We then use a {\it questioning strategy} to determine which element of $\mathcal{Q}$ should be asked to the MuE. In particular, we propose a novel strategy that selects the element that maximally reduces the uncertainty in the performance model (Sec.~\ref{sec:QuestioningStrategy}). The process terminates after a fixed number of questions have been asked or when the uncertainty in the model has been reduced to an acceptable level. Fig.~\ref{fig:GPexamples} (Left) illustrates our framework and we detail our performance model and questioning strategy next.

\subsection{Performance Model}
\label{sec:perfModel}
From a set of questions and the corresponding MuE responses, we aim to model the MuE performance with respect to concepts $\mathcal{C}$. While the relation between concepts could in practice be complex, we model them as independent here.  

By its definition, the MuE provides answers to binary questions, and for any concept, there are 4 possible outcomes to a question: a True Negative (TN), a False Positive (FP), a False Negative (FN) or a True Positive (TP). 
\begin{figure}[!t]
	\centering
	\includegraphics[width=0.99\textwidth]{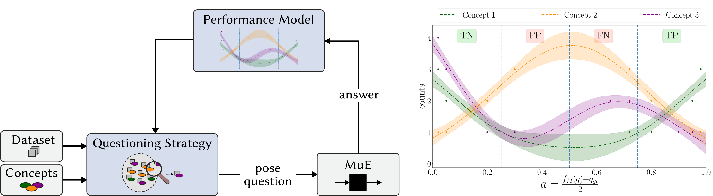}
	\caption{{{\bf Left:} Overview of the proposed scheme. {\bf Right:} Examples of GPs for 3 different concepts (points are the observations, dotted line is the mean of a GP and the shaded area corresponds to the $95\%$ confidence region).}}
	\label{fig:GPexamples}
\end{figure}
Our goal then is to model the relation between the frequency of these outcomes and the inputs to the MuE. To do this, we define a discrete random variable $Y_c$ for every concept $c \in \mathcal{C}$ which encodes the counts\footnote{The approach would be unchanged if the outcome frequency was used.} of outcomes given by $f$. We achieve this by means of a Gaussian Process (GP)~\cite{Rasmussen06gaussianprocesses} of the form, 
\begin{equation}
f_{c}^{\mathcal{GP}}(a_c) \sim \mathcal{GP}\left(\mu_c (a_c), k_c(a_c, a_c')\right),
\end{equation}
\noindent
where $\mu_c(\cdot)$ and $k_c(\cdot,\cdot)$ are the mean and kernel functions of the GP, respectively, and
\begin{equation}
a_c = \frac{f(q) + q_{gt}}{2},
\end{equation}
\noindent 
describes the answer of $f$ with respect to the question $q$. The mean function $\mu_c$ is initialized to $0$, and the covariance function (kernel) $k_c$ is a squared exponential $k_c (a_{c,m}, a_{c,n}) = \sigma_f \cdot e^{-\frac{1}{2l^2}(a_{c,m} - a_{c,n})^2} + \sigma_n \delta_{mn}$, with characteristic length-scale $l=0.1$, initial signal variance $\sigma_f=1$ and noise variance $\sigma_n=0.025$.

To then infer the value of $Y_c$ for any $a_c$, we store the pairs $\{(a^{(i)}_{c},y_{c}^{(i)})\}$ where $y_{c}^{(i)}$ is the number of times $f$ has given $a=a^{(i)}_{c}$, and use these as observations to infer the complete model $f_c^{\mathcal{GP}}$ using standard inference~\cite{Rasmussen06gaussianprocesses}. In practice, we discretized the range of $a$ in bins of $\Delta a = 0.01$.

A consequence of this model is that we can now visualize the performance of $f$ with respect to the concepts in $\mathcal{C}$, the selected $q$'s and the dataset $\mathcal{D}$. In Fig.~\ref{fig:GPexamples} (Right), we illustrate $f^{\mathcal{GP}}_c$ in terms of $a$ for each $c$. Such a visualization depicts any bias that both $f$ and the dataset $\mathcal{D}$ may have (\eg $\mathcal{D}$ contains few samples regarding a specific concept). Note that by summing up the observations over concepts or integrating over the four different subregions of the support set, one retrieves the total TN, FP, FN and TP, and other subsequent metrics.
  
\begin{algorithm}[t!]
	\caption{Concept Centric VTT}\label{alg:questioning_strategy}
\begin{algorithmic}
\Require Dataset: $\mathcal{D}$, Concepts: $\mathcal{C}$, stopping criteria: $\tau$ and MuE: $f$
\State $\mathcal{Q \gets D\times C}$
\State Initialize $f^{\mathcal{GP}}_c, c\in\mathcal{C}$
\While{ \# questions asked $< \tau$}
\State $\mathcal{Q}_{candidates} \gets \left\{ q \in \mathcal{Q} : c_q \in \argmax\limits_{c \in \mathcal{C}}{ \left( \max \left(u_c^-, u_c^+ \right) \right)} \text{ and } q_{gt}=o_{c_q} \right\}$
\If {select based only on uncertainty}
\State $q^* \gets \text{ randomly select } q\in \mathcal{Q}_{candidates}$
\ElsIf {select based on uncertainty \& unpredictability}
\State $q^* \gets \text{ random } q\in \{q \in \mathcal{Q}_{candidates}: |p(f(q)=\text{\textquotedblleft Yes\textquotedblright}|\mathcal{H} ) - 0.5| < \epsilon  \}$
\EndIf
\State Compute $f(q^*)$, $a_c$
\State Update $f^{\mathcal{GP}}_c$
\State $\mathcal{H} \gets \{ \mathcal{H}, (q^*, f(q^*) \}$
\State $\mathcal{Q} \leftarrow  \{ \mathcal{Q} \setminus q^* \}$
\EndWhile
\end{algorithmic}
\end{algorithm}

\subsection{Questioning Strategy}
\label{sec:QuestioningStrategy}
With the performance model above and the fact that the validation dataset  $\mathcal{D}$ may be intractably larger, we now describe how to select samples from $\mathcal{Q}$ to verify that the MuE has grasped relevant concepts.  

To do this, we present a strategy that looks to select samples $\mathbf{s}$ and concepts $c$ that are likely to reduce our model's uncertainty. We do this by computing the uncertainty of a concept as the integral of the $95\%$ confidence region of the GP over its support set (\ie 2 standard deviations). That is, for concept $c \in C$, 
\begin{equation}
u_c = 4 \int_0^1{k_c(a,a)da},
\label{eq:GPUnc}
\end{equation}
\noindent
which can be decomposed as $u_c^- = 4\int_0^{0.5}{ k_c(a,a)da}$ and $u_c^+ = 4\int_{0.5}^1{k_c(a,a)da}$, corresponding to the uncertainty associated with negative and positive samples in $\mathcal{D}$, respectively. Visually this corresponds to the area over the intervals $a = [0,0.5]$ and $a=[0.5,1]$ (see Fig.~\ref{fig:GPexamples}, Right).

Our strategy then chooses which concept to ask, and whether to ask a sample that has or does not have this concept in it. This is performed by selecting
\begin{equation}
q^* \in \left\{ q \in \mathcal{Q} : c_q \in \argmax_{c \in \mathcal{C}}{(\max(u_c^-, u_c^+))} \text{ and } q_{gt}=o_{c_q} \right \}, 
\label{eq:uncertainty_candidates}
\end{equation}
\noindent 
where $o_{c_q} = 0$ if $\max (u_c^- , u_c^+) = u_c^- $ or 1 if $\max(u_c^-, u_c^+) = u_c^+$. 

From this, $q^*$ is selected either randomly, or based on its unpredictability. As in~\cite{geman2015visual}, the latter is computed using the same dataset $\mathcal{D}$ and the history of already answered questions $\mathcal{H}$. An overview of the questioning strategy can be seen in Algorithm~\ref{alg:questioning_strategy}.

\section{Experiments and Results}
We choose to evaluate our framework on multi-label classification tasks where concepts are directly linked to groundtruth labels. In the following, we outline the different datasets and MuEs we use, and our experimental setup.

\subsection{Datasets and MuE}

{\bf Indian Diabetic Retinopathy image Dataset (IDRiD)~\cite{h25w98-18}:} {143 fundus images} from both healthy and diabetic retinopathy subjects, with the task of identifying 4 different lesion types (hemorrhage, hard and soft exudates, microaneurysm). The multilabel MuE is a ResNet~\cite{he2016deep} with pre-trained weights, trained with 100 images. The remaining 43 images were used as samples in $\mathcal{D}$. 

\noindent {\bf ISIC 2018 Skin Lesion Analysis\footnote{\url{https://challenge2018.isic-archive.com/}}:} 1'876 dermoscopic images of skin lesions. Here $N_C=5$, consisting of different skin lesions types: pigment network, negative network, globules, milia like cysts and streaks. The MuE is a pre-trained ResNet~\cite{he2016deep} trained with 70\% of data. A hundred randomly chosen images from the 30\% test  images were used to populate $\mathcal{D}$. 
	
\noindent {\bf OCT:} 200 OCT cross-sectional slices from Age-Related Macular Degeneration and Diabetic Macular Edema patients. Eleven different biomarkers potentially can be present in any cross-section ($N_C=11$). All 200 images are used for $\mathcal{D}$ and a seperate trained Dilated Residual Network~\cite{Yu2017-gu} is used for the MuE.

Given that we are not focused on optimizing the performance of a specific MuE but rather on evaluating relative behavior with respect to concepts, we also provide a set of synthetically generated MuEs for which we understand their performance fully. That is, given that the distribution of concepts in $\mathcal{D}$ is not uniform for any of the datasets, we wish to compare each MuE to a \textquotedblleft biased\textquotedblright ~MuE. To do this, we simulate a biased MuE that answers with $90\%$ accuracy questions regarding the most common concept and $50\%$ on all others. Similarly, we simulate a MuE with a $50\%$ accuracy for the most common concept, and $90\%$ on all others. Last, we also simulate an unbiased algorithm, with a $70\%$ accuracy regardless of the concept or class imbalance.

\subsection{Experiments}
To evaluate our framework, we compare four questioning strategies: (a) random, (b) based on the predictability of the question given the previous questions and the dataset~\cite{geman2015visual}, (c) based on the uncertainty of the question as defined in Sec.~\ref{sec:QuestioningStrategy}, and (d) based on the combination of the above two, as described in Sec.~\ref{sec:QuestioningStrategy}.
\begin{figure}[!t]
	\centering
	\subfloat[\small{random}]{\includegraphics[width=0.48\textwidth]{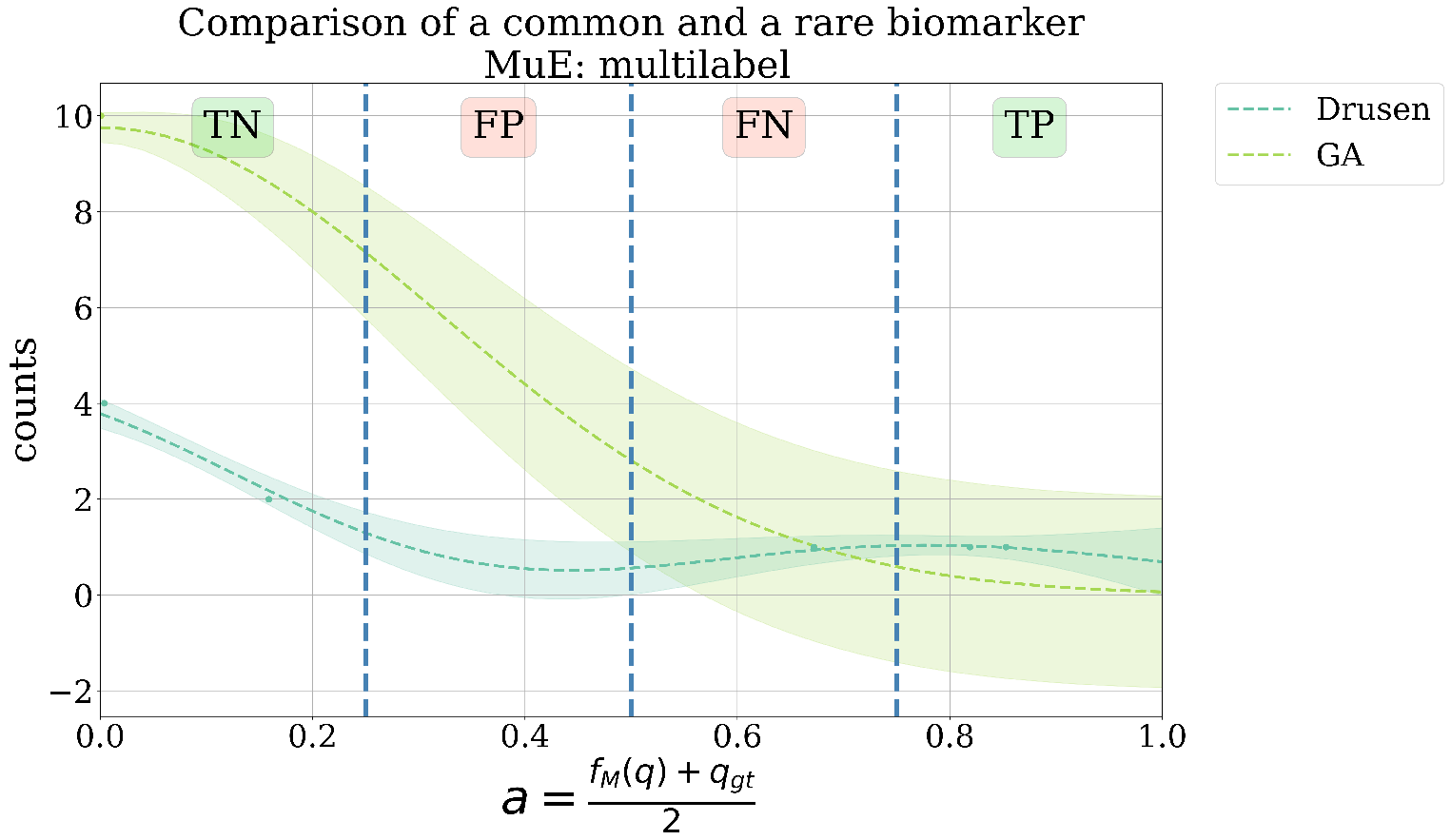}}
	\subfloat[\small{unpredictability}]{\includegraphics[width=0.48\textwidth]{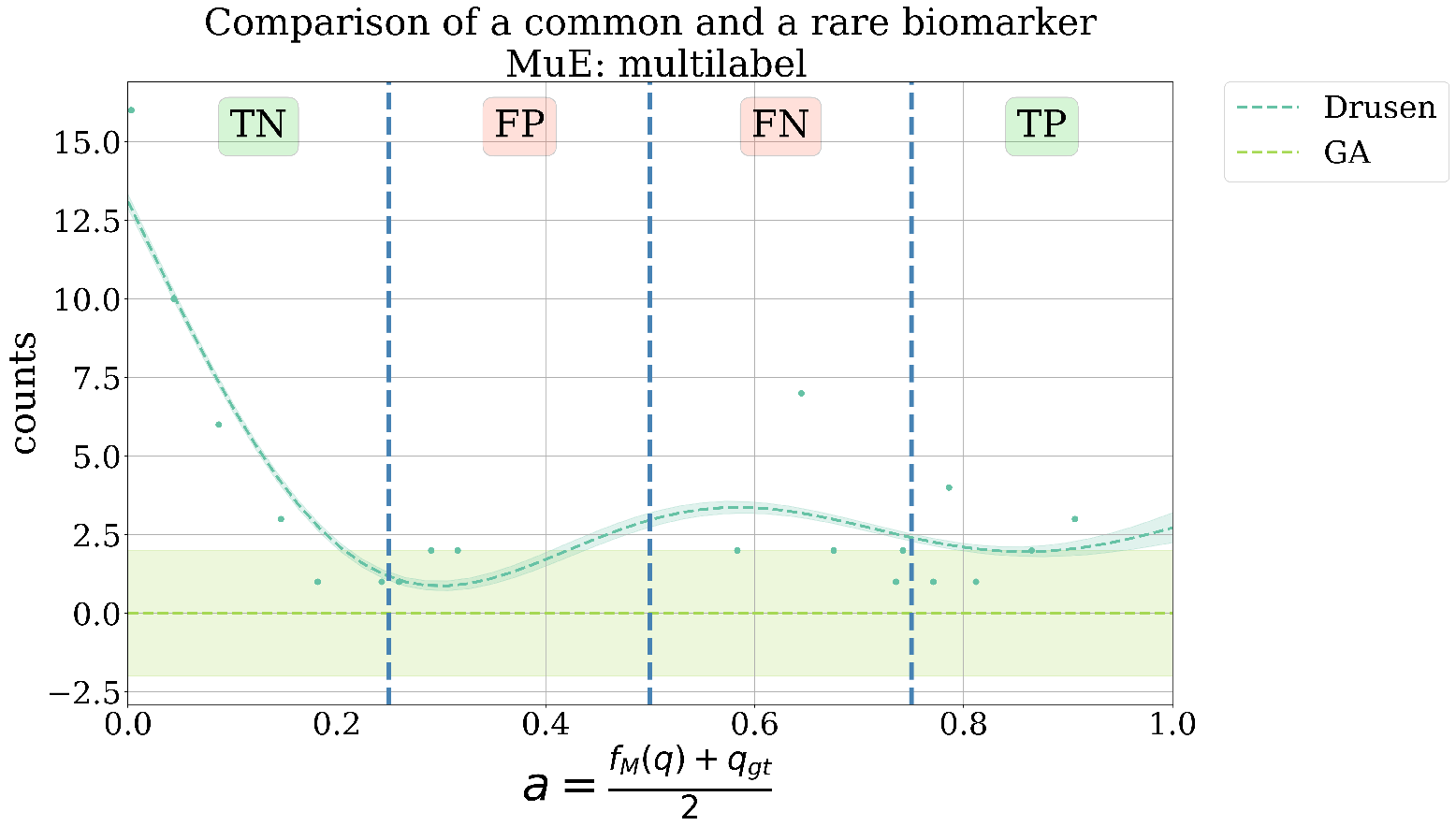}}
	\\
	\subfloat[\small{uncertainty}]{\includegraphics[width=0.48\textwidth]{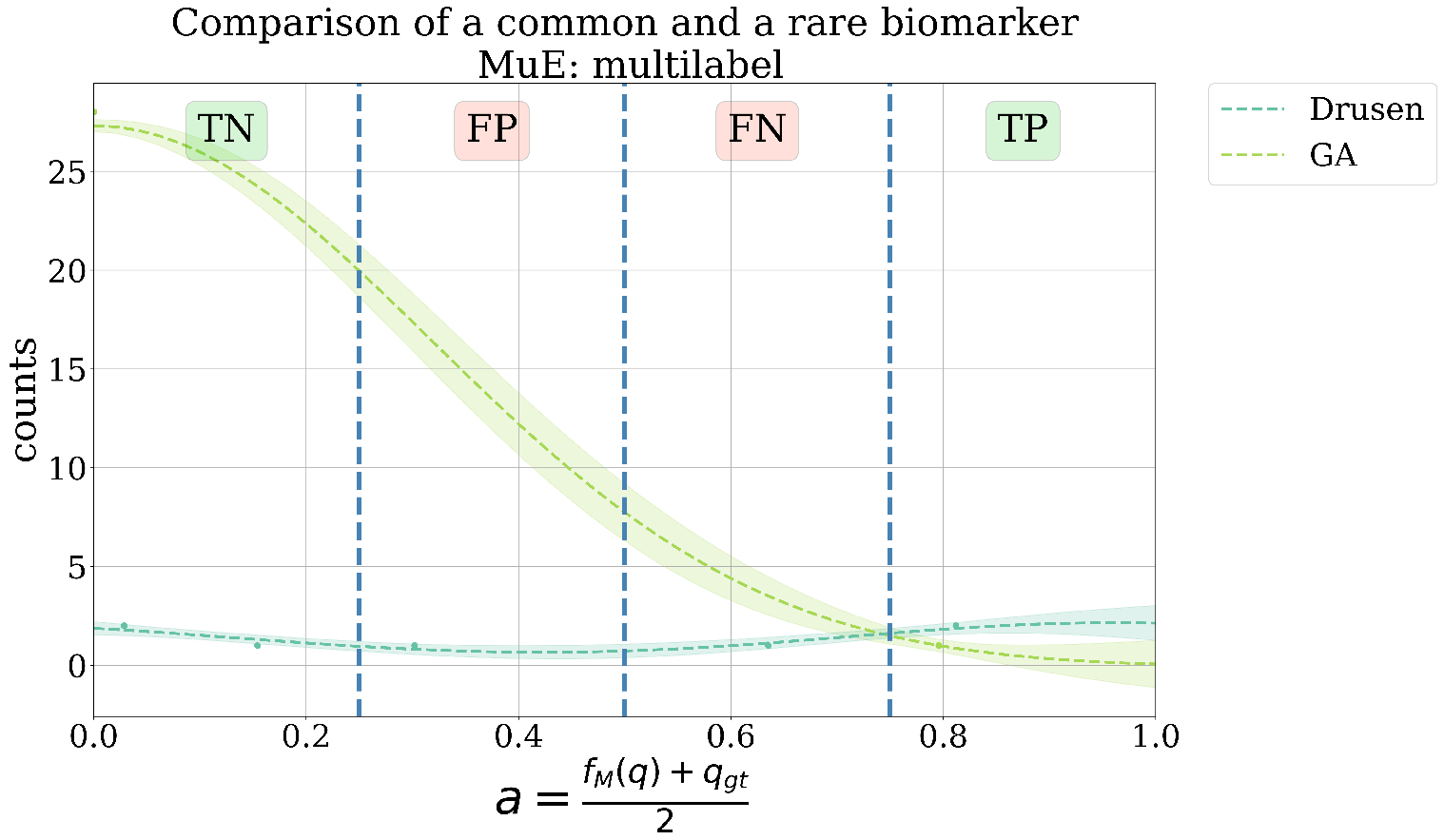}}
	\subfloat[\small{uncertainty \& unpredictability}]{\includegraphics[width=0.48\textwidth]{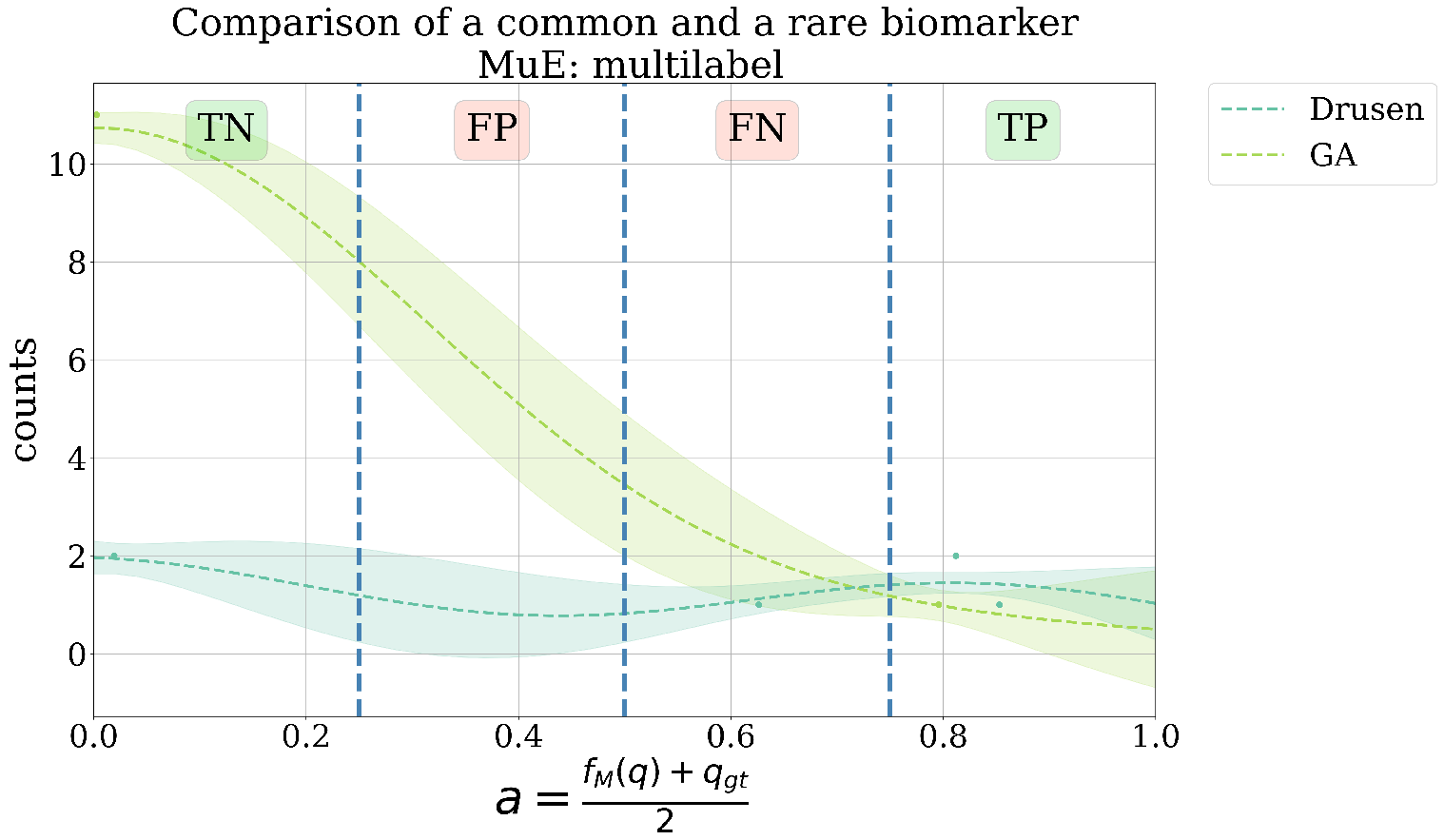}}
	\caption{{Performance model for the 4 questioning strategies. The most common concept (Drusen) and a rare concept (Geographic Atrophy - GA) are shown after 100 questions are posed to the multilabel MuE.}}
	\label{fig:rare_dominant}
\end{figure}

\begin{figure}[!t]
	\centering
	\includegraphics[width=\textwidth]{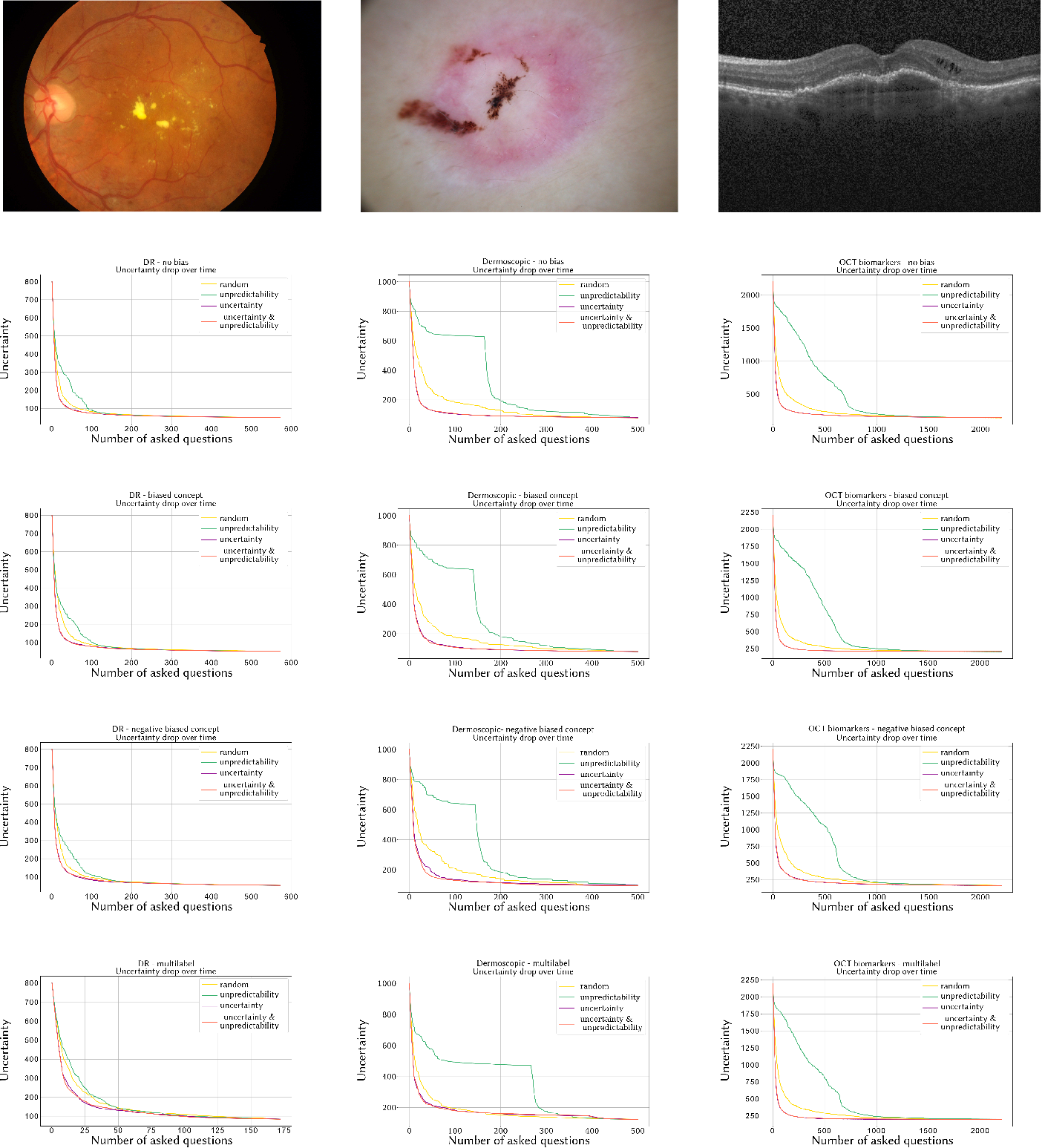}
	\caption{{Uncertainty of the performance model with respect to the number of  questions asked for different MuEs. Each column corresponds to a dataset. \textbf{Rows, from top to bottom:} image example, uncertainty for unbiased MuE, uncertainty for biased to most common concept MuE, uncertainty for negatively biased to most common concept MuE and uncertainty for trained multilabel classifier.}}
	\label{fig:results_all}
\end{figure}

Fig.~\ref{fig:rare_dominant} depicts the state of the performance for two concepts (GA and Drusen) after a hundred questions have been asked on the~{\bf OCT} dataset using the trained MuE. In all plots the bias in concept occurrence can be observed and we see that both the random strategy, and the one relying solely on the unpredictability criteria, are prone to not ask many questions about rare concepts (GA concept), thus delaying the assessment of the MuE's on this concept. Both uncertainty based methods however manage to sample adequately and are more confident in the performance model.

In another experiment, we ask all possible questions following each one of the four described strategies. That is, the state of the performance model is the same after all questions have been asked. We repeat the experiments 10 times, and monitor the average uncertainty as questions are asked. The results are shown in Fig.~\ref{fig:results_all}. Here we observe that the proposed questioning strategies reach saturated levels of uncertainty quicker than the other strategies, and would require fewer questions for confident assessments.
\section{Discussion}
To summarize, we present a more informative and interpretable performance model for evaluating closed-end, \textquotedblleft Yes/No\textquotedblright~ inference methods. To this, we propose a strategy to sample from the - possibly intractable - set of questions, in order to reach high certainty in the performance model characterizing the MuE and the validation data. We assess our method on three different medical imaging datasets and show that the performance model is able to capture the data distribution information and the MuE biases. Moreover, the questioning strategy allows for faster convergence of the performance model to a low uncertainty state. We will look to extend this concept to segmentation problems in future work.

\bibliographystyle{splncs04}
\bibliography{refs} 
\end{document}